\documentclass{article} 
\usepackage{iclr2016_workshop,times}
\usepackage{hyperref}
\usepackage{url}
\usepackage{graphicx}
\usepackage{bm,amsmath}
\usepackage{array}
\newcolumntype{P}[1]{>{\centering\arraybackslash}p{#1}}

\title{Neural network-based clustering using pairwise constraints}

\author{Yen-Chang Hsu \\
School of Electrical and Computer Engineering\\
Georgia Institute of Technology\\
Atlanta, GA 30332, USA \\
\texttt{yenchang.hsu@gatech.edu} \\
\And
Zsolt Kira \\
Georgia Tech Research Institute \\
Atlanta, GA 30318, USA \\
\texttt{zkira@gatech.edu} \\
}

\begin{document}

\maketitle

\begin{abstract}
This paper presents a neural network-based end-to-end clustering framework. We design a novel strategy to utilize the contrastive criteria for pushing data-forming clusters directly from raw data, in addition to learning a feature embedding suitable for such clustering. The network is trained with weak labels, specifically partial pairwise relationships between data instances. The cluster assignments and their probabilities are then obtained at the output layer by feed-forwarding the data. The framework has the interesting characteristic that no cluster centers need to be explicitly specified, thus the resulting cluster distribution is purely data-driven and no distance metrics need to be predefined. The experiments show that the proposed approach beats the conventional two-stage method (feature embedding with k-means) by a significant margin. It also compares favorably to the performance of the standard cross entropy loss for classification. Robustness analysis also shows that the method is largely insensitive to the number of clusters. Specifically, we show that the number of dominant clusters is close to the true number of clusters even when a large $k$ is used for clustering. 
\end{abstract}

\section{Introduction}

Performing end-to-end training and testing using deep neural networks to solve various tasks has become a dominant approach across many fields due to its performance, efficiency, and simplicity. Success across a diverse set of tasks has been  achieved in this manner, including classification of pixel-level information into high level categories ~\citep{Alex2012imagenet}, pixel-level labeling for image segmentation ~\citep{long2015segmentation,zheng2015conditional}, robot arm control ~\citep{levine2015end}, speech recognition ~\citep{graves2014towards}, playing Atari games ~\citep{mnih2015human} and Go ~\citep{clark2015training,David2016go}. All of the above techniques largely avoid sophisticated pipeline implementations and human-in-the-loop tuning by adopting the concept of training the networks to learn the target problem directly.

Clustering, a classical machine learning problem, has not yet been fully explored in a similar manner. Although there are some two-stage approaches that have tried to learn the feature embedding specifically for clustering, they still require using other clustering algorithms such as k-means to determine the actual clusters at the second step. Specifically, the first stage of previous works usually assume how the data is distributed in the projected space using human-chosen criteria such as self-reconstruction, local relationship preservation, sparsity ~\citep{tian2014learning,huang2014deep,shao2015deep,wang2015learning,chen2015deep,song2013auto}, fitting predefined distributions ~\citep{xie2015unsupervised}, or strengthening of neighborhood relationships ~\citep{rippel2015metric} to learn the feature embedding. Furthermore, all of these techniques then use a metric, such as Euclidean or cosine distance, in the second stage. This further introduces human-induced bias via strong assumptions and the chosen metric may not necessarily be appropriate for the embedded space. In other words, there has not been a method to solve the two sub-problems (learning a feature space and performing clustering within that feature space) jointly in an end-to-end manner.

In this work, we propose a framework which minimizes such assumptions by training a network that can directly assign the clusters at the output layer. We specifically use weak labels, in the form of pairwise constraints or similar/dis-similar pairs, to learn the feature space as well as output a clustering. It is worth emphasizing that such weak labels could be obtained automatically (in an unsupervised manner) based on spatial or temporal relationships, or using a neighborhood assumption in the feature space similar to the above works. One could also get the weak labels from the ground-truth obtained from crowd-sourcing. In many cases, it may be an easier task for a human to provide pair-wise relationships rather than direct assignment of class labels (e.g. when dealing with attribute learning). 

In order to adopt the raw data and weak labels for end-to-end clustering, we present the novel concept of constructing the cost function in a manner that incorporates contrastive KL divergence to minimize the statistical distance between predicted cluster probabilities for similar pairs, while maximizing the distance for dissimilar pairs. In the latter sections, we will show that the framework is extremely easy to realize by rearranging existing functional blocks of deep neural networks, so it has large flexibility to adopt new layer types, network architectures, or optimization strategies for the purpose of clustering.

One significant property of the proposed end-to-end clustering is that there are no cluster centers explicitly represented. This largely differs from all of the works mentioned above. Without the centers, no explicit distance metrics need to be involved for deciding the cluster assignment. The learning of the cluster assignments is purely data-driven and is implicitly handled by the parameters and the non-linear operations of the network. Of course the outputs of the last hidden layer could be regarded as the learned features, however it is not necessary to interpret it using predefined metrics such as Euclidean or cosine distance. The networks will find the best way to utilize the embedded feature space during the same training process in order to perform clustering. The experimental sections will demonstrate this property, in addition to strong robustness when the number of output clusters is varied. In such cases, the network tends to output a clustering that only utilizes the same number of nodes as there are clusters intrinsically in the data.

Furthermore, since the proposed framework can learn the cluster assignments using the proposed contrastive loss, it opens up the possibility of directly comparing its accuracy to the standard cross-entropy loss when full labels are available. This is achieved by developing an implementation that can efficiently utilize such dense pairwise information. This  implementation strategy and experiments are also presented in sections \ref{impl} and \ref{cvc}, showing favorable results compared to the standard classification approach. Source code in Torch is provided on-line.

\subsection{Related Works}

A common strategy to utilize pairwise relationship with neural networks is the Siamese architecture \citep{bromley93}. The concept had been widely applied to various computer vision topics, such as similarity metric learning \citep{chopra2005learning}, dimensionality reduction \citep{hadsell2006dimensionality}, semi-supervised embedding \citep{weston2008deep}, and some applications to image data, such as in learning to match patches \citep{han2015matchnet,Zagoruyko_2015_CVPR} and feature points \citep{simo2015discriminative}. The work of \citet{mobahi2009deep} uses the coherence nature of video as a way to collect the pairwise relationship and learn its features with a Siamese architecture. The similar idea of leveraging temporal data is also presented in the report of \citet{goroshin2015unsupervised}. In addition, the triplet networks, which could be regarded as an extension of Siamese, gained significant success in the application of learning fine-grained image similarity \citep{wang2014learning} and face recognition \citep{schroff2015facenet}. Despite the wide applicability of the Siamese architecture, there is no report exploring them from the clustering perspective. Furthermore, while some works try to maximize the information in a training batch by carefully sampling the pair \citep{han2015matchnet} or by formulating it as a triplet \citep{wang2014learning}, there is no work showing how to use dense pairwise information directly and efficiently.

Our proposed implementation strategy can efficiently utilize any amount of pairwise constraints from a dataset to train a neural network to perform clustering. When the full set of constraints is given, it can compare to the vanilla networks trained using supervised classification. If only partial pairwise constraints are available, the problem is similar to semi-supervised clustering. There is a long list of previous work related to the problem. For example, COP-Kmeans \citep{wagstaff01} forced the clusters to comply with constraints and \citet{rangapuram12} added terms in spectral clustering to penalize the violation of constraints. The more closely related works perform metric learning \citep{bilenko04} or feature re-weighting \citep{minkowski12} during the clustering process. The recent approaches TVClust and RDP-means \citep{khashabi2015clustering} address the problem with probabilistic models. None of these approaches, however, jointly learn the feature space in addition to clustering. 

In the next sections we will explain how to inject the concept of clusters into a neural network formulations. The experiments on two image datasets will be presented in the third section, demonstrating the efficacy of the approach.

\section{The End-to-End Clustering Networks}
\label{gen_inst}

Consider the vanilla multilayer perceptron (MLP) used for classification tasks: Each output node is associated with predefined labels and the optimization minimizes a cost function, such as cross entropy, that compares the output labels (or the distribution over the labels) provided by the network for a set of instances and the corresponding ground truth labels. We start from this model and remove the hard association between labels and network outputs. The idea is to only use pairwise information and define the output nodes in a manner such that they can represent a clustering of the data. In other words, which node will correspond to which cluster (or object class) is dynamically decided during the training process. To achieve this, we formulate an approach that only needs to modify the cost criterion above the softmax layer of any neural network which was designed for a classification task. We therefore present a new pairwise cost function for clustering that can take the place of, or be combined with, the traditional supervised classification loss functions. This flexibility allows the network to use both types of information, depending on which is available.

\subsection{Pairwise KL-Divergence}

\begin{figure*}
	\centering
	\includegraphics[width=0.8\linewidth]{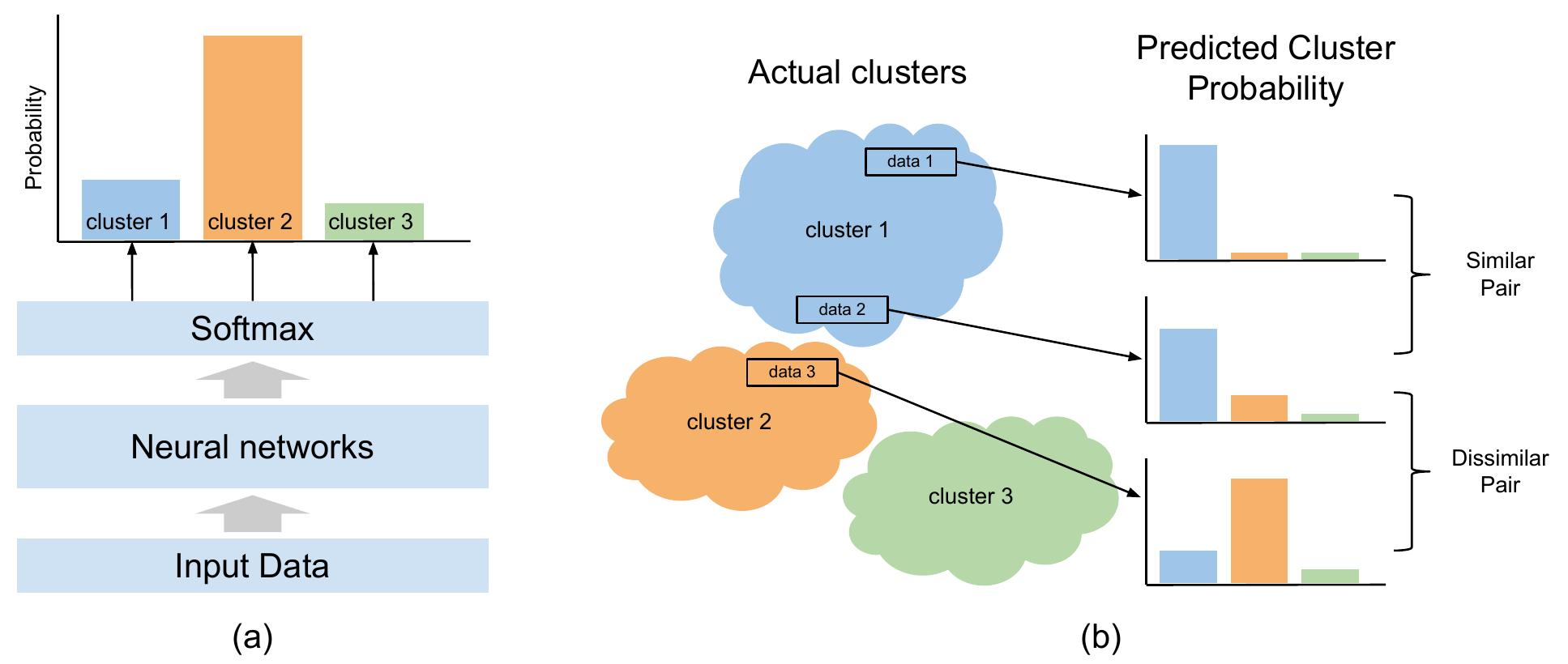}
	\caption{Illustration of (a) how neural networks output the distribution of possible clusters given a sample, (b) the example of predicted cluster distribution between similar/dissimilar pairs.}
	\label{fig:concept_diagram}
\end{figure*}

While the output of the traditional softmax layer represents the probability that a sample belongs to the class labels (or clusters in our problem),  the outputs of the whole softmax layer could be viewed as the distribution of possible clusters given a sample (Figure \ref{fig:concept_diagram}a). If the data only contains a single concept, such as in the case of hand-written digits, then the distributions between the softmax output for a similar pair should be similar. Conversely, the distribution over the class labels should be dissimilar if the pair belongs to different clusters (Figure \ref{fig:concept_diagram}b). The similarity between distributions could be evaluated by statistical distance such as Kullback-Leibler (KL) divergence. Traditionally this can be used to measure the distance between the output distribution and ground truth distribution. In our case, however, it can instead be used to measure the distance between the two output distributions given a pair of instances. Given a pair of distributions $\bm{P}$ and $\bm{Q}$, obtained by feeding data $x_p$ and $x_q$ into network $f$, we will fix $\bm{P}$ first and calculate the divergence of $\bm{Q}$ from $\bm{P}$. Assume the network has $k$ output nodes, then the total divergence will be the sum over $k$ nodes. To turn the divergence into a cost, we define that if $\bm{P}$ and $\bm{Q}$ come from a similar pair, the cost will be plain KL-divergence; otherwise, it will be the hinge loss (still using divergence). The indicator functions $I_s$ in equation \ref{eq2} will be equal to one when $(x_p,x_q)$ is a similar pair, while $I_{ds}$ works in reverse manner. In other words:

\[\bm{P}=f(x_p), \bm{Q}=f(x_q),\]
\begin{equation}
KL(\bm{P} \parallel \bm{Q})=\sum_{i=1}^k{P_i log(\frac{P_i}{Q_i})},
\end{equation}

\begin{equation}\label{eq2}
loss(\bm{P} \parallel \bm{Q})=I_s(x_p,x_q)KL(\bm{P} \parallel \bm{Q})+I_{ds}(x_p,x_q)\max{(0,margin-KL(\bm{P} \parallel \bm{Q}))}.
\end{equation}

Since the cost should be calculated from fixing both $P$ or $Q$ (i.e. symmetric), the total cost $L$ of the pair $x_p$, $x_q$ is the sum of both directions:
\begin{equation}\label{eq3}
L(\bm{P},\bm{Q})=loss(\bm{P} \parallel \bm{Q})+loss(\bm{Q} \parallel \bm{P}).
\end{equation}

To calculate the derivative of cost $L$, it is worth to note that the $P$ in the first term of equation \ref{eq3} (and $Q$ in the second term) is regarded as constant instead of variable. Thus, the derivative could be formulated as:
\begin{equation}\label{eq4}
\begin{split}
\frac{\partial}{\partial Q_i}L(\bm{P},\bm{Q})&=\frac{\partial}{\partial Q_i}loss(\bm{P} \parallel \bm{Q}), \\
&=\left\{\begin{array}{ccc}-\frac{P_i}{Q_i} & \mbox{if} & I_s(x_p,x_q)=1, \\
\frac{P_i}{Q_i} & \mbox{elseif} & KL(\bm{P} \parallel \bm{Q}))<margin, \\
0 & \mbox{otherwise}. & \end{array}\right.
\end{split}
\end{equation}
\begin{equation}\label{eq5}
\begin{split}
\frac{\partial}{\partial P_i}L(\bm{P},\bm{Q})&=\frac{\partial}{\partial P_i}loss(\bm{Q} \parallel \bm{P}), \\
&=\left\{\begin{array}{ccc}-\frac{Q_i}{P_i} & \mbox{if} & I_s(x_p,x_q)=1, \\
\frac{Q_i}{P_i} & \mbox{elseif} & KL(\bm{Q} \parallel \bm{P}))<margin, \\
0 & \mbox{otherwise}. & \end{array}\right.
\end{split}
\end{equation}

With the defined derivatives of cost, the standard back-propagation algorithm can be applied without change.

\subsection{Efficient implementation to Utilize Pairwise Constraints}
\label{impl}

\begin{figure}
	\centering
	\includegraphics[width=1\linewidth]{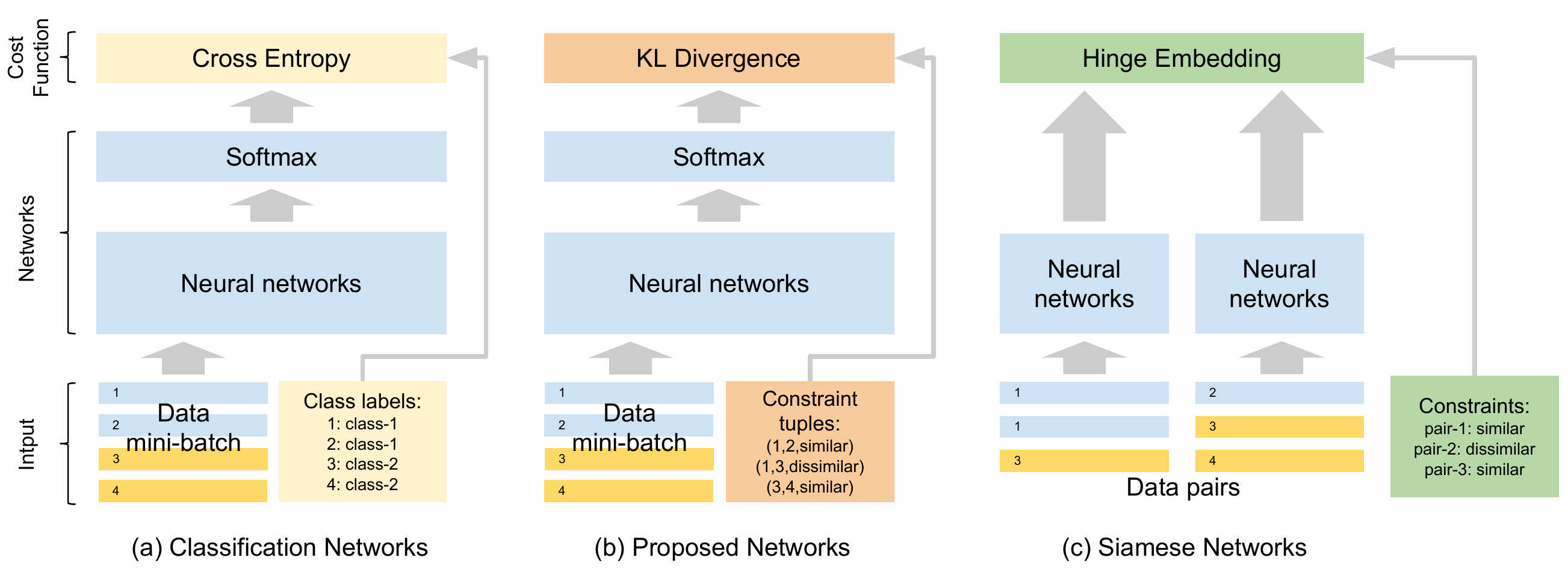}
	\caption{The comparison between (a) classification networks, (b) our proposed networks, and (c) Siamese networks. The parts that differ across architectures are shown with distinct colors. In (a) and (b), the numbers in the data represent the index of the input data in a mini-batch.}
	\label{fig:architecture_diagram}
\end{figure}

Equation \ref{eq2} is in the form of contrastive loss that is suitable to be trained with Siamese networks \citep{hadsell2006dimensionality}. However, when the amount of pairwise constraints increases, it is not efficient to enumerate all pairs of data and feed them into Siamese networks. Specifically, if there is a mini-batch that has pairwise constraints between any two samples, the number of pairs that have to be fed into the networks will be $n(n-1)/2$ where $n$ is mini-batch size. However, a redundancy occurs when a sample has more than one constraint associated with it. In such cases the sample will be fed-forward multiple times. However, feed-forward once for each sample is sufficient for calculating the pairwise cost in a mini-batch. Figure \ref{fig:architecture_diagram}c demonstrates an example for the described situation. The data with index 1 and 3 are fed-forward twice in vanilla Siamese networks to enumerate the three pairwise relationships: (1,2),(1,3), and (3,4). To avoid the redundancy of computation, we apply a strategy of enumerating the pairwise relationships only in the cost layer, instead of instantiating the Siamese architecture. This strategy simplified the implementation of neural networks which utilize pairwise relationship. Our proposed architecture is shown in the Figure \ref{fig:architecture_diagram}b. The pairwise constraints only need to be presented to the cost layer in the format of tuples $T:(i,j,relationship)$ where $i$ and $j$ are the index of sample inside the mini-batch and $relationship$ indicates similar/dissimilar pair. Each input data is therefore only fed-forward once in a mini-batch and its full/partial pairwise relationships are enumerated as tuples.

Concretely, the gradients for the back-propagation in a mini-batch are calculated as:
\begin{equation}
	\frac{\partial}{\partial f(x_i)}\hat{L}=\sum_{\forall{j};(i,j)\in{T}}\frac{\partial}{\partial f(x_i)}L(f(x_i),f(x_j)).
\end{equation}

One could see our proposed architecture (Figure \ref{fig:architecture_diagram}b) is highly similar to the standard classification networks (Figure \ref{fig:architecture_diagram}a). As a result of this design, ideas in the above two sections could be easily implemented as a cost criterion in the torch {\em nn} module. Then a network could be switched to either classification mode or clustering mode by simply changing the cost criterion. We therefore implemented our approach in Torch, and have released the source on-line \footnote{http://github.com/yenchanghsu/NNclustering}.

It is also worth mentioning that the presented implementation trick is not specifically for the designed cost function. Any contrastive loss could benefit from the approach. Since there is no openly available implementation to address this aspect, we include it in our released demo source.

\section{Experiments}

We evaluate the proposed approach on the MNIST \citep{lecun98} and CIFAR-10 \citep{krizhevsky09} datasets. The two datasets are both normalized to zero mean and unit variance. The convolutional neural networks architecture used in these experiments is similar to LeNet \citep{lecun98}. The network has 20 and 50 5x5 filters for its two convolution layers with batch normalization \citep{ioffe2015batch} and 2x2 max-pooling layers. We use the same number of filters for both MNIST and CIFAR-10 experiments. The two subsequent fully connected layers have 500 and 10 nodes. Both convolutional and the first fully connected layers are followed by rectified linear units. The only hyper-parameter in our cost function is the $margin$ in equation \ref{eq2}. The margin was chosen by cross-validation on the training set. There is no significant difference when the margin was set to 1 or 2. However, it has a higher chance of converging to a lower training error when the margin is 2, thus we set it to the latter value across the experiments. To minimize the cost function, we applied mini-batch stochastic gradient descent.

\subsection{Clustering with partial constraints}

We performed three sets of experiments to evaluate our approach. The first experiment seeks to demonstrate how the approach works with partial constraints. In this case, we use a clustering metric to demonstrate how good the resulting clustering is. The constraints are uniformly sampled from the full set, i.e, ${\#full\textendash{constraints}=n(n-1)/2}$, where $n$ is the size of training set. The pairwise relationship is converted from the class label. If a pair has the same class label, then it is a similar pair, otherwise it is dissimilar. We did not address the fact that the amount of dissimilar pairs usually dominates the pairwise relationship (which is more realistic in many application domains), especially when the number of classes is large. In our experiments for this section, the ratio between the number of similar and dissimilar pairs is roughly 1:9.

We evaluate the resulting clusters with the purity measure and normalized mutual information (NMI) \citep{strehl2003cluster}. The index of cluster for each sample is obtained by feed-forwarding the training/testing data into the trained networks. Note that we collect the clustering results of training data after the training error has converged, i.e, feed the training data one more time to collect the outputs after the training phase. We picked the networks which have the lowest training loss among five random restarts while the set of constraints are kept the same.  

\begin{figure}
	\centering
	\includegraphics[width=0.9\linewidth]{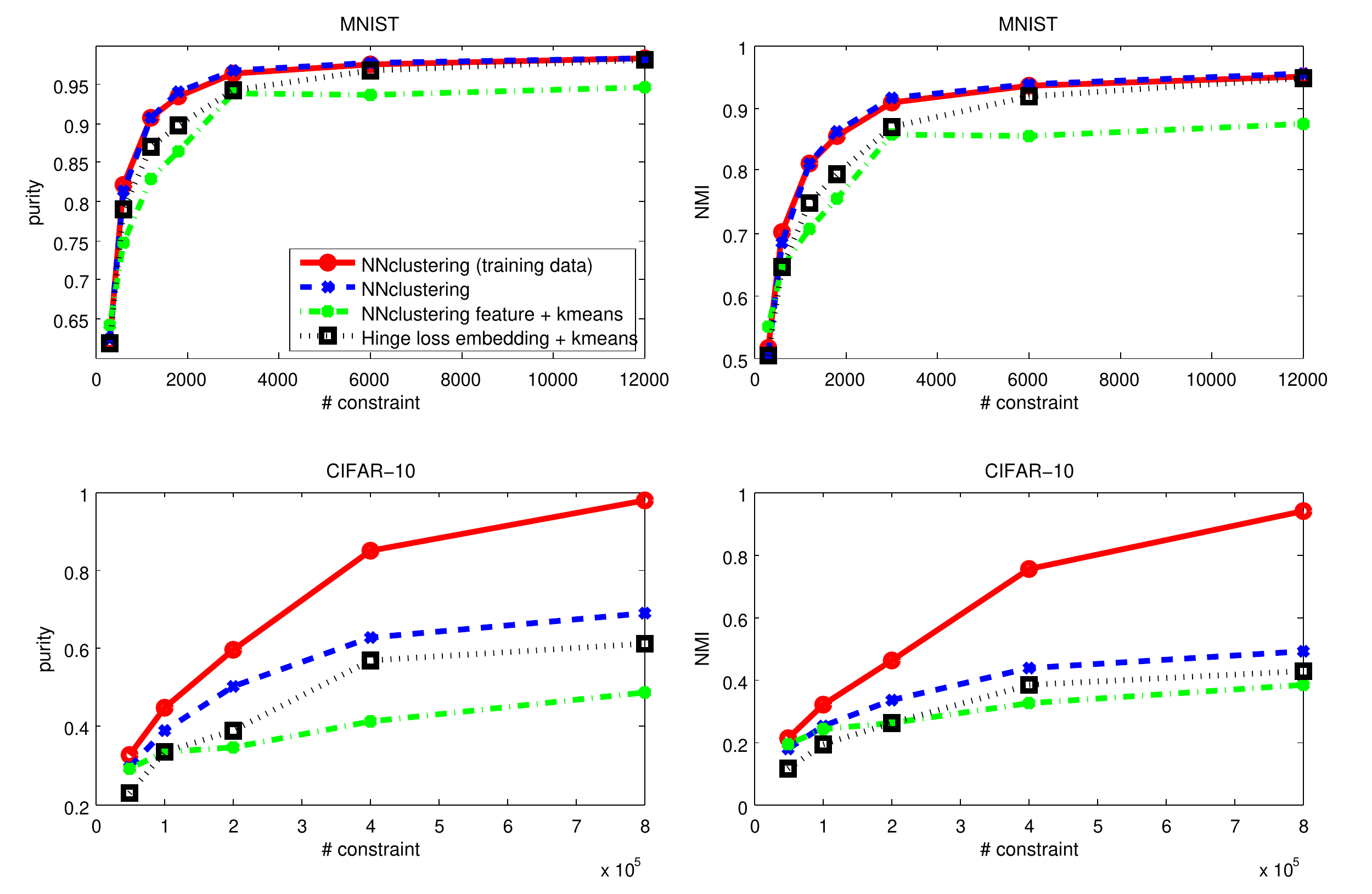}
	\caption{The results of clustering with partial pairwise constraints. The \#constraint axis is the number of sampled pairwise relationship in the training data. The clustering and training is simultaneously applied on the training data (red line). The testing data (for blue, green, black lines) is used to validate if the feature space learned during clustering has generalizability. NNclustering is our proposed method. The \textit{NNclustering feature + kmeans} uses the outputs at the last hidden layer (500-D) as the input for k-means. The baseline networks (black line) were trained with hinge loss of Euclidean distance. The evaluation metric in the first column is purity, while the second column shows the NMI score.}
	\label{fig:cluster_result}
\end{figure}

Figure \ref{fig:cluster_result} shows that on MNIST the clustering could still achieve high accuracy when constraints are extremely sparse. With merely 1200 constraints, which were randomly sampled from the pairwise relationship of full (60000 samples) training set, it achieves \textgreater0.9 purity and \textgreater0.8 NMI scores. Note that the training samples without any constraint associated to it has no contribution to the training. Thus, the scheme is not the same as the semi-supervised clustering framework in previous works \citep{wagstaff01,bilenko04,minkowski12} where their unlabeled data contribute to calculating the centers of the clusters. The lack of explicit cluster centers provides the flexibility to learn more complex non-linear representations, so the proposed algorithm could still predict the cluster of unseen data without knowing the cluster centers. In the experiments with MNIST, we could see the performance of testing data has no degradation. It is mainly because the networks could learn the clustering with so few constraints such that most of the training data have no constraints and act like unseen data.  Note that although no directly comparable results for MNIST have been reported for our specific problem formulation, results for the closest problem setting can be seen in Figure 4(b) of \citet{li2009constrained} which achieves similar results for a subset of the classes and a much smaller number of constraints (only about 3k). Hence, our results are competitive with theirs but our approach is scalable enough to allow the use of many more constraints and continues to improve while their approach seems to plateau. 

\begin{figure}
	\centering
	\includegraphics[width=0.8\linewidth]{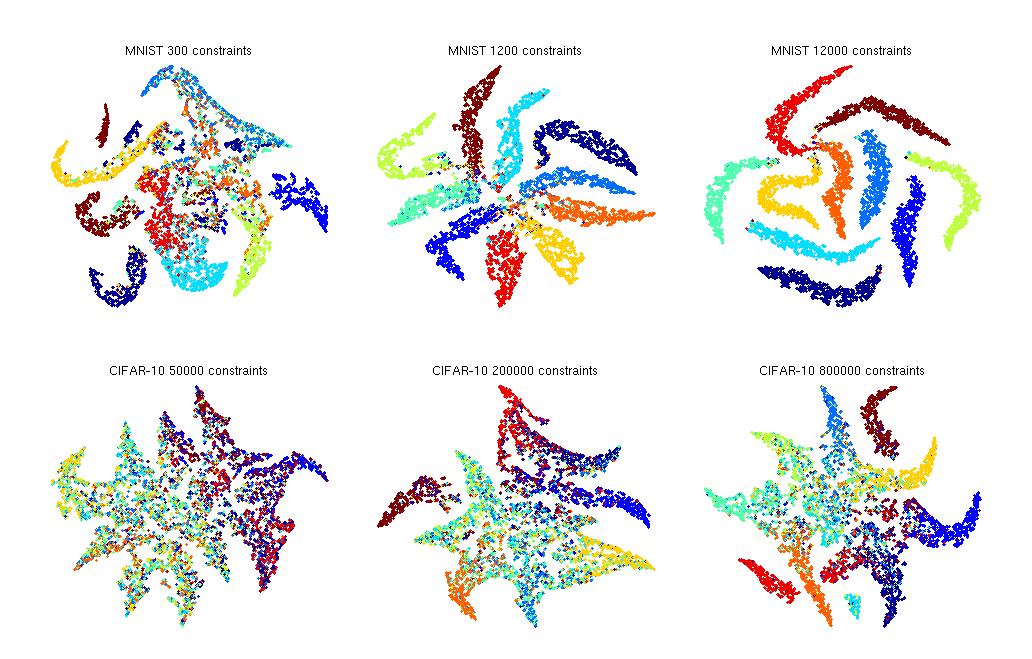}
	\caption{The visualization of clustering. The figure was created by using the outputs of the softmax layer as the input for t-SNE \citep{van2008visualizing}. Only testing data are shown. The networks used in the first row are trained with 300, 1200, and 12000 pairs of constraints in MNIST training set. The second row is trained with 50k, 200k, and 800k constraints in CIFAR-10.}
	\label{fig:vis_clustering}
\end{figure}

To demonstrate the advantage of performing joint clustering and feature learning, we also applied the k-means algorithm with the features learned at the last hidden layer, which has 500 dimensions. The k-means algorithm used Euclidean or cosine distance and was deployed with 50 random restarts on the testing set. We report the clustering results of k=10 which has the lowest sum of point-to-centroid distances among 50 restarts. Since the dimensionality is relatively high, the performance of using Euclidean and cosine distance showed minor difference. The results in Figure \ref{fig:cluster_result} show that the jointly trained last layer utilize the outputs of last hidden layer much better than k-means.

To construct the baseline approach, we use the common strategy of training a Siamese networks with standard hinge loss embedding criteria in \textit{torch nn} package, then perform k-means on the networks' outputs. The baseline networks have the same architecture except the softmax layer and the loss function. Figure \ref{fig:cluster_result} shows that the proposed clustering framework beats the baseline with a significant margin when the number of constraints is few in the easy dataset (purity is ${\sim}5\%$ better in MNIST) or when the dataset is harder (purity is $15{\sim}50\%$ better in CIFAR-10).

The experiments with CIFAR-10 provides some idea of how the approach works on a more difficult dataset. The required constraints to achieve reasonable clustering is much higher. Eight constraints/sample (400,000 total constraints) is required to reach a 0.8 purity score with the same network. The performance on unseen data is also degraded because the networks is over-fitting the constraints. The degradation could possibly be mitigated by adding some regularization terms such as dropout. While any general regularization strategy could be applied in the proposed scheme, we do not address it in this work. Nevertheless, the clustering on the training set is still effective with sparse constraints, e.g., it is able to reach a purity of ${\sim}1$ with only 16 constraints/sample on CIFAR-10. The visualization in Figure \ref{fig:vis_clustering} provides more intuition about the clustering results trained with different numbers of constraints.

\subsection{Robustness of Clustering} 

\subsubsection{Adding Noise} 
Noisy constraints are likely to occur when the pairwise relationships are generated in an automatic/unsupervised way. We simulated this scenario by flipping the sampled pairwise relationship. Since the ratio of similar pair and dissimilar pair is 1:9, adding 10\% noise will introduce equal amount of false-similar pair as the amount of true-similar pair. The clustering performance in Figure  \ref{fig:robustness} (left) shows the reasonable tolerance against noise. We would like to point out that when noise is less than 10\%, the performance degradation is reduced when the number of constraints increased. This means that the proposed method could achieve higher performance by adding more pairwise information while keeping the ratio of noise the same. Real applications would benefit from this property since adding more weakly labeled data is cheap and the noise level of automatically generated constraints are usually the same.

\subsubsection{Changing Number of Clusters} 

\begin{figure}
	\centering
	\includegraphics[width=1\linewidth]{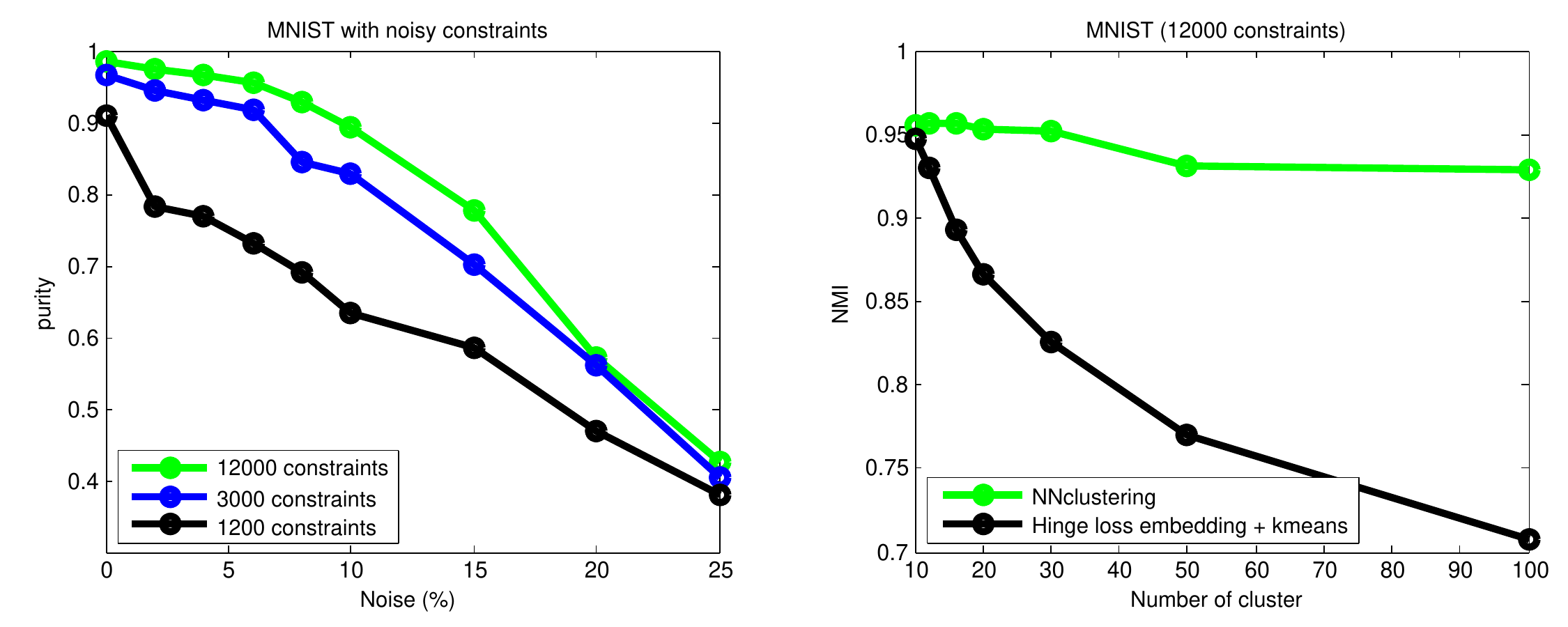}
	\caption{The robustness evaluation of the proposed clustering method. Left figure is the result of adding noisy constraints into MNIST, while the right figure simulates the case when the number of clusters is unknown.}
	\label{fig:robustness}
\end{figure}

\begin{figure}
	\centering
	\includegraphics[width=0.6\linewidth]{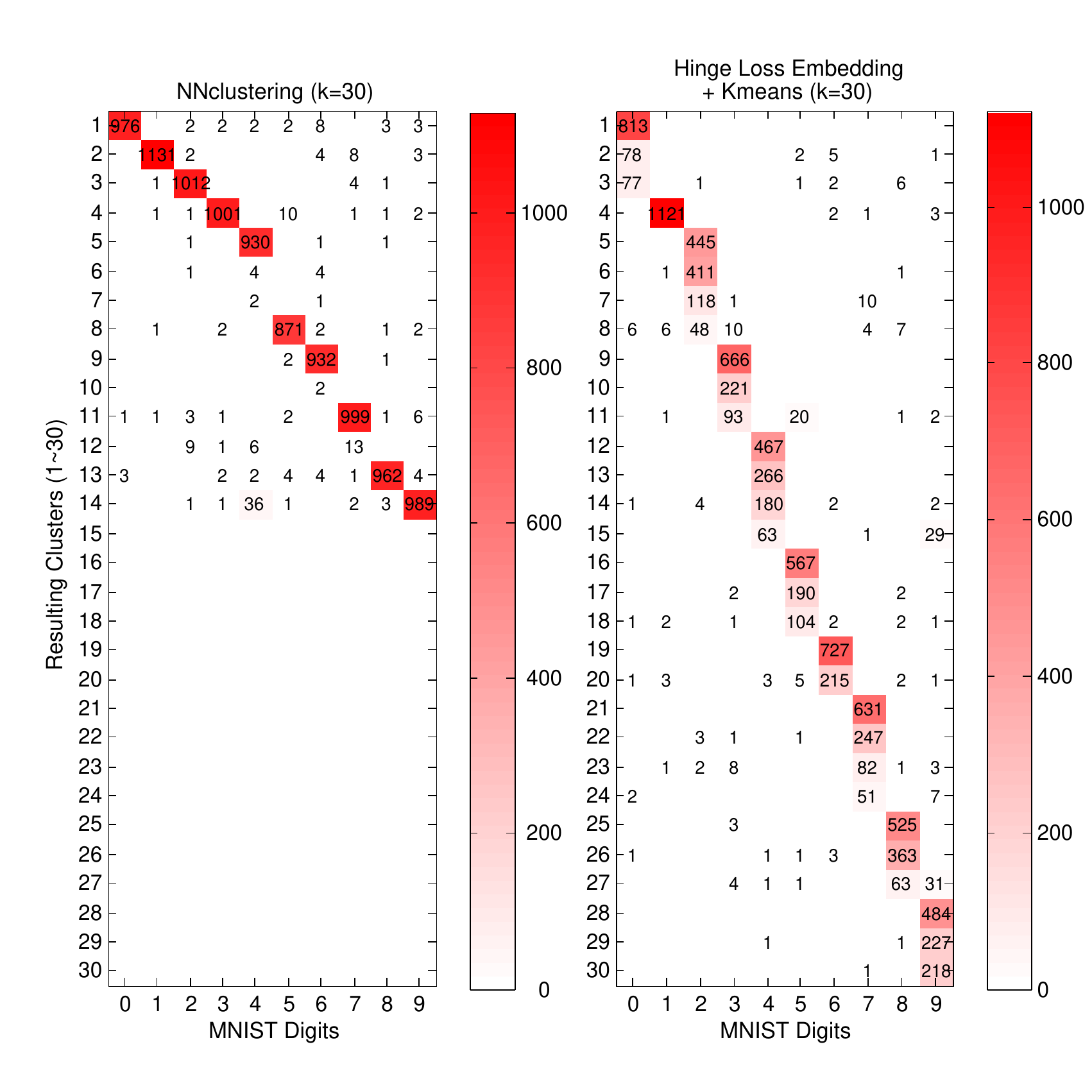}
	\caption{The contingency tables of resulting clusters. It only shows k=30 (same experiment in the right part of figure \ref{fig:robustness}) for the ease of visualization. NNclustering produces similar result even when k=100. The numbers in the table show the amount of samples been assigned to the cluster, while the blank rows indicate empty clusters. Higher numbers in fewer positions is the preferred result for clustering.}
	\label{fig:confusion}
\end{figure}

Another common scenario is that the number of target clusters is unknown. Since the purity metric is not sensitive to the number of clusters, NMI is more appropriate in this evaluation. We performed the experiment with 12,000 constraints for training, which include ${\approx}$1200 similar pairs in MNIST. The testing results in the right of figure \ref{fig:robustness} show that the proposed method is almost not affected by increasing the number of clusters. Even in the condition of 100 clusters (by setting 100 output nodes in our networks), the performance only decreases by a very small amount. In fact, in figure \ref{fig:confusion} most of the data were assigned to $\approx$10 major clusters and left other clusters being empty. In contrast, the kmeans-based approach (hinge loss embedding + kmeans) is susceptible to the number of clusters and usually divide a class into many small clusters.

\subsection{Clustering VS Classification} \label{cvc}

The final set of experiments compares the accuracy of our approach with a pure classification task in order to get an upper bound of performance (since full labels can be used to create a full set of constraints) and see whether our approach can leverage pairwise constraints to achieve similar results. To make the results of clustering (contrastive loss) comparable to classification (cross-entropy loss), the label of each cluster is obtained from the training set. Specifically,  we make the number of output nodes to be the same as the true number of classes, thus we could assign each output node with a distinct label using the optimal assignment. The results in Table \ref{table:cls_vs_clu} show our cost function achieved slightly higher or comparable accuracy in most of the experiment settings. The exception is MNIST with 6 samples/class. The reason is that the proposed cost function creates more local minimum. If the training data is too few, then the training will be more likely to be trapped in certain local minimum. Note that we also applied a random restart strategy (randomly initializing the parameters of the network) to find a better clustering result based on the training set, which is a common strategy used in typical clustering procedures. We ran 5 randomly initialized networks to perform clustering and chose the network that had the highest training accuracy and then used the resulting network to predict the clusters on the testing set. 

We also performed the experiments using the same architecture applied to a harder dataset, i.e, CIFAR-10. We did not pursue optimal performance on the dataset, but instead used it to compare the performance difference of learning between the classification and clustering networks. The results show that they are fully comparable. Since CIFAR-10 is a much more difficult dataset compared to MNIST, the overall drop of accuracy on CIFAR-10 is reasonable. Even in the extreme case when the number of training samples is small, the proposed architecture and cost function proved effective.

\begin{table}	
	\caption{Comparing the testing accuracy between classification and clustering using same networks architecture. The clustering is trained with full pairwise relationships obtained from ground-truth class labels. The separated testing set (10,000 samples) is used in this evaluation.}
	\begin{center}
		\begin{tabular}{|P{5cm}|P{2cm}|P{2cm}|}
			\hline	
			Training approach & Classification & Clustering  \\ 
			\hline	
			Training data: & & \\
			MNIST 6 sample/class	& \textbf{82.4\%} & 79.4\% \\ 
			MNIST 60 sample/class	& 94.7\%          & \textbf{95.1\%} \\ 
			MNIST 600 sample/class	& 98.3\%          & \textbf{98.8\%} \\ 
			MNIST full ($\approx$6000 sample/class)	& 99.4\%  & \textbf{99.6\%} \\ 
			\hline
			Training data: & & \\	
			CIFAR-10 5 sample/class	    & 21.3\%          & \textbf{22.0\%} \\ 
			CIFAR-10 50 sample/class	& 34.6\%          & \textbf{37.0\%} \\ 
			CIFAR-10 500 sample/class	& \textbf{55.0\%} & 53.2\% \\ 
			CIFAR-10 full (5000 sample/class)	& \textbf{73.7\%}  & 73.4\% \\
			\hline	
		\end{tabular}
	\end{center}
	\label{table:cls_vs_clu}
\end{table}

\section{Conclusion and Future Works}
\label{headings}

We introduce a novel framework and construct a cost function for training neural networks to both learn the underlying features while, at the same time, clustering the data in the resulting feature space. The approach supports both supervised training with full pairwise constraints or semi-supervised with only partial constraints. We show strong results compared to traditional K-means clustering, even when it is applied to a feature space learned by a Siamese network. Our robustness analysis not only shows good tolerance to noise, but also demonstrates the significant advantage of our method when the number of clusters is unknown. We also demonstrate that, using only pairwise constraints, we can achieve equal or slightly better results than when explicit labels are available and a classification criterion is used. In addition, our approach is both easy to implement for existing classification networks (since the modifications are in the cost layer) and can be efficiently implemented. 

In future work, we plan to deploy the approach using deeper network architectures on datasets that have a larger number of classes and instances. We hope that this work inspires additional investigation into feature learning via clustering, which has been relatively less explored. Given the abundance of available data and recent emphasis on semi or unsupervised learning as a result, we believe this area holds promise for analyzing and understanding data in a manner that is flexible to the available amount and type of labeling.  

\subsubsection*{Acknowledgments}

This work was supported by the National Science Foundation and National Robotics Initiative (grant \# IIS-1426998).

\bibliography{iclr2016_NNclustering}
\bibliographystyle{iclr2016_workshop}

\end{document}